\pdfoutput=1
\documentclass[11pt]{article}

\usepackage[preprint]{acl}

\usepackage{times}
\usepackage{latexsym}

\usepackage[T1]{fontenc}
\usepackage[utf8]{inputenc}
\usepackage{CJKutf8}
\usepackage{microtype}
\usepackage{graphicx}
\usepackage{inconsolata}

\usepackage{times}
\usepackage{subfigure}
\usepackage{microtype}
\usepackage{amssymb}
\usepackage{amsmath}
\usepackage{multirow}
\usepackage{subfigure}
\usepackage{xcolor}
\usepackage{colortbl}
\usepackage{makecell}
\usepackage{multirow}
\title{AC-EVAL: Evaluating Ancient Chinese Language Understanding in Large Language Models}

 \author{Yuting Wei, Yuanxing Xu, Xinru Wei, Simin Yang, Yangfu Zhu, \\ {\bf Yuqing Li, Di Liu, Bin Wu}\thanks{Corresponding Author.} \\ Beijing Key Laboratory of Intelligence Telecommunication Software and Multimedia, \\Beijing University of Posts and Telecommunications\\ \{yuting\_wei, xyx, wxr2000, ysm2000, zhuyangfu, liyuqing, liudi, wubin\}@bupt.edu.cn}
\begin{document}
\maketitle
\begin{CJK}{UTF8}{gbsn}
\begin{abstract}
Given the importance of ancient Chinese in capturing the essence of rich historical and cultural heritage, the rapid advancements in Large Language Models (LLMs) necessitate benchmarks that can effectively evaluate their understanding of ancient contexts. To meet this need, we present AC-EVAL, an innovative benchmark designed to assess the advanced knowledge and reasoning capabilities of LLMs within the context of ancient Chinese. AC-EVAL is structured across three levels of difficulty reflecting different facets of language comprehension: general historical knowledge, short text understanding, and long text comprehension. The benchmark comprises 13 tasks, spanning historical facts, geography, social customs, art, philosophy, classical poetry and prose, providing a comprehensive assessment framework. Our extensive evaluation of top-performing LLMs, tailored for both English and Chinese, reveals a substantial potential for enhancing ancient text comprehension. By highlighting the strengths and weaknesses of LLMs, AC-EVAL aims to promote their development and application forward in the realms of ancient Chinese language education and scholarly research.\footnote{The AC-EVAL data and evaluation code are available at \url{https://github.com/yuting-wei/AC-EVAL}.}
\end{abstract}

\section{Introduction}

The advent of Large Language Models (LLMs) has significantly impacted Natural Language Processing (NLP), highlighting their importance in understanding and generating human languages \cite{wei2022emergent, zhou2022least, zhao2023survey}. With the rise of Chinese as a major global language, there has been a surge in Chinese-specific LLMs \cite{zeng2022glm, bai2023qwen, baichuan2023baichuan2}. Ancient Chinese, a crucial part of the Chinese language, records a rich historical and cultural heritage, and has garnered considerable attention from computational linguists \cite{li-etal-2022-multi-modal, wang-etal-2023-enhancing}.
LLMs present significant opportunities for enhancing the pedagogy of Chinese literary education through convenient text analysis and comprehension.
Therefore, assessing the ancient Chinese comprehension capabilities of LLMs holds significant importance.

Initially, benchmarks for LLMs primarily targeted the assessment of English language understanding, exemplified by MMLU \cite{hendrycks2021measuring}, BIG-bench \cite{srivastava2023beyond} and HELM \cite{liang2023holistic}. Subsequently, several benchmarks focusing on Chinese, such as C-Eval \cite{huang2023ceval}, CMMLU \cite{li2023cmmlu}, and SuperCLUE \cite{xu2023superclue}, were introduced. These benchmarks aim to evaluate the reasoning performance of LLMs across a broad spectrum of fields including STEM, social sciences, and humanities. However, these benchmarks tend to lean towards modern Chinese comprehension. While some include tasks related to Chinese language, literature and history, they are often relegated to minor categories, insufficient for a comprehensive coverage of ancient Chinese knowledge and language assessment.
Existing benchmarks for ancient Chinese understanding, such as CCLUE \footnote{https://github.com/Ethan-yt/guwen-models} and WYWEB \cite{zhou-etal-2023-wyweb}, cover various aspects but primarily focus on linguistic feature analysis, frequently overlooking the assessment of historical knowledge hidden in literature. Furthermore, the diversity in format across these datasets, tailored for specific tasks rather than providing a unified assessment framework, complicates the evaluation of LLMs, presenting challenges in conducting uniform assessments.

\begin{figure*}[htbp]
	\centering
	\includegraphics[width=1\linewidth]{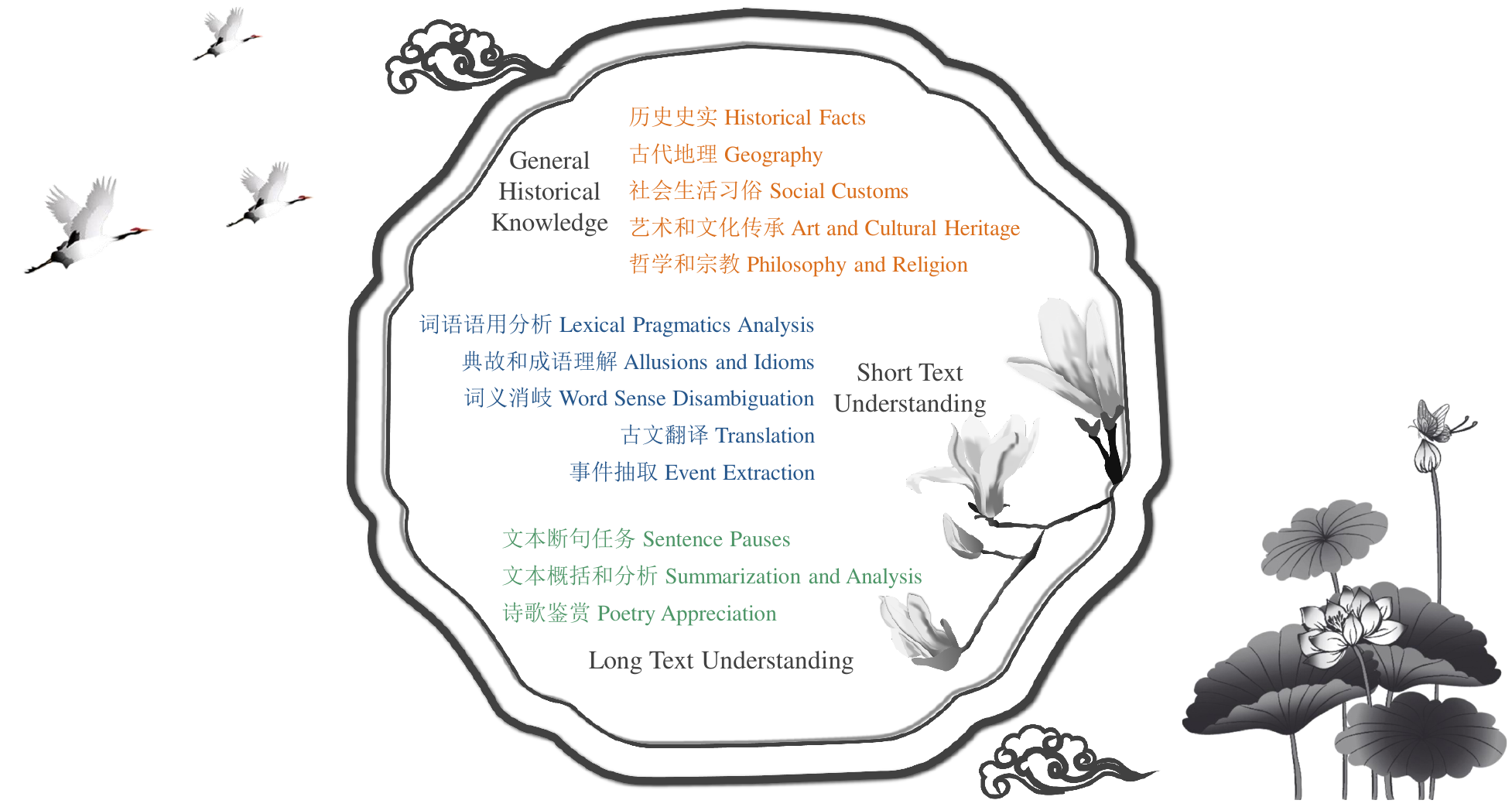}
	\caption{Overview of AC-EVAL.}
	\label{overview}
\end{figure*}

To bridge this gap, we propose AC-EVAL (as illustrated in Figure \ref{overview}), a benchmark meticulously designed for a comprehensive evaluation of LLMs' proficiency in ancient Chinese language understanding and historical knowledge.  AC-EVAL comprises 3,245 multiple-choice questions, spanning three distinct dimensions and thirteen subjects, covering historical periods from the Pre-Qin to the Qing dynasty. These tasks, which progressively increase in difficulty, are categorized into general historical knowledge, short text understanding, and long text understanding.
The general historical knowledge tasks address a diverse range of contents, including but not limited to, ancient historical facts, geography, social customs, art, religion and philosophy.  Short text understanding covers lexical semantics and pragmatics, allusions and idioms, sentence translations, and event extraction. Long text understanding tasks focus on long text pauses, classical prose summarization and analysis, and the appreciation of themes, emotions and styles in poetry.

In our evaluation of LLMs on the AC-EVAL benchmark across answer-only (AO) and chain-of-thought (COT) settings in zero- and few-shot scenarios, only ERNIE-Bot 4.0 and GLM-4 with accuracies over 70\%. Results reveal significant improvement potential, especially in long text comprehension.
Our analysis shows that Chinese LLMs outperform English ones in ancient Chinese. This distinction underscores the unique challenge that ancient Chinese as a low-resource area for models like GPT-4, despite their commendable performance on other Chinese benchmarks.
Moreover, the broad range of knowledge required in our tasks reveals that LLMs encounter difficulties in grasping underlying rules, affecting few-shot learning outcomes.
Interestingly, zero-shot COT shows an advantage in larger models, underscoring the value of reasoning steps for complex tasks. 
Through the AC-EVAL benchmark, our goal is to provide a multidimensional evaluation tool, highlighting potential improvement areas to advance the development of LLMs in the understanding and education of ancient Chinese.

\section{Related Work}

\subsection{Chinese benchmarks for LLMs}

In the evolving landscape of NLP, the development of benchmarks to evaluate LLMs in comprehending Chinese has been a focal point of recent research \cite{chang2023survey}.
Benchmarks such as MMCU \cite{zeng2023measuring}, C-Eval \cite{huang2023ceval}, and CMMLU \cite{li2023cmmlu} derived primarily from official examination questions, spanning various disciplines including STEM, humanities, social sciences, and professional qualification tests for fields like law and medicine. These benchmarks aimed to comprehensively assess the breadth of domains relevant to the Chinese language, primarily utilizing multiple-choice questions as their core components.
AGIEval \cite{zhong2023agieval} expanded upon these by incorporating fill-in-the-blank questions alongside multiple-choice.
CG-Eval \cite{zeng2023evaluating} and CLEVA \cite{li2023cleva}, on the other hand, took a more holistic approach to measure models' generative abilities, including tasks such as noun explanation, short answer questions, and computational problems.
SuperCLUE \cite{xu2023superclue} evaluated models across three dimensions: foundational abilities, professional knowledge, and Chinese language characteristics by leveraging actual user queries and ratings, along with a mix of open- and closed-ended questions.
Lastly, OpenCompass \cite{2023opencompass} integrates over 100 public datasets into a unified leaderboard framework, standardizing the assessment of LLMs.

Despite the extensive range of current benchmarks, there is a significant gap in their coverage of the ancient Chinese language, literature, and history. 
Considering the depth and breadth of Chinese millennia-long history, which includes evolving social customs, religious beliefs, geographical boundaries, and linguistic changes, it is evident that a more comprehensive benchmark is necessary. 

\subsection{Ancient Chinese benchmarks}

Ancient Chinese, a fundamental component of the Chinese linguistic heritage, encapsulates millennia of historical narratives and cultural wisdom. A multitude of traditional and diverse datasets has been proposed to evaluate the ancient Chinese language understanding capabilities with various specific tasks \cite{pan-etal-2022-zuo, wang-ren-2022-uncertainty, liu2022contrastive, tang-su-2022-slepen}.

For instance, analyzing the sentiments and themes in poetry (as seen in FSPC \cite{fspc} and TCCP \cite{yutong2020construction}, to the intricate task of translating between classical and modern Chinese, (illustrated by the Classical-Modern corpus\footnote{https://github.com/NiuTrans/Classical-Modern} and the Erya dataset \cite{guo2023towards}). Furthermore, named entity recognition and relationship extraction tasks, with datasets like C-CLUE \cite{ji2021c} and GuNER 2023\footnote{https://guner2023.pkudh.org}, provide a foundation for in-depth linguistic analysis within ancient texts.
GuwenEE\footnote{https://github.com/Lyn4ever29/GuwenEE}, an event extraction dataset,
is annotated and constructed from the "Twenty-Four Histories," a collection of Chinese official historical literature.
The word sense disambiguation dataset for ancient Chinese, introduced by \citet{shu-etal-2021-gu}, encompasses texts from multiple dynasties.
Additionally, the EvaHan series from 2022 to 2024 introduces a spectrum of tasks including sentence segmentation, POS tagging, and machine translation.
Comprehensive benchmarks like CCLUE and WYWEB \cite{zhou-etal-2023-wyweb} integrate a variety of language understanding tasks, ranging from text classification to poetry analysis and machine reading comprehension, offering a holistic evaluation of models’ linguistic proficiency.

However, despite the breadth of these benchmarks, there remains a discernible gap in the assessment of models' grasp of the historical knowledge hidden within ancient texts. The varied formats of datasets, designed for specific tasks, hinder uniform LLM evaluation, highlighting the urgent need for an integrated benchmark to thoroughly assess LLMs' understanding of ancient Chinese literature and history knowledge.

\section{AC-EVAL Overview}

\begin{table*}[h!]
	\centering
	\begin{tabular}{lcccc}
		\hline
		\textbf{Category} & \textbf{Difficulty} & \textbf{\# Subjects} & \textbf{\# Questions} & \textbf{Average Length}\\ \hline
		General Historical Knowledge & Easy & 5 & 1014 & 62.78\\ 
		Short Text Understanding & Normal & 5 & 1215 & 214.19\\ 
		Long Text Understanding  & Hard & 3 & 1016 & 536.95 \\ \hline
		%		\multicolumn{4}{c}{\itshape In terms of split} \\ 
		%		Test & 52 & 3191 & -\\ 
		%		Dev & 52 & 65 & -\\ \hline
		%		Total & 52 & 3256 & -\\ \hline
	\end{tabular}
	\caption{Statistics of AC-EVAL. The average length is measured in characters.}
	\label{stat}
\end{table*}

\subsection{Design Principles}
The motivation behind constructing AC-EVAL is to comprehensively assess LLMs' understanding and reasoning capabilities regarding the shifts in societal customs, culture, and language throughout millennia of history. 
It adheres to four foundational principles to ensure a holistic evaluation framework:

\textbf{Temporal Coverage:} It spans from the pre-Qin period to the Qing dynasty, offering a wide historical scope that covers thousands of years of evolution.

\textbf{Task Difficulty Diversity:} The benchmark ranges from basic fragmented historical knowledge to complex tasks requiring the understanding of ancient Chinese texts of various lengths, providing a graded evaluation of model capabilities.

\textbf{Content Diversity:} It encompasses a broad spectrum of knowledge areas including historical facts, geography, religion, philosophy, social customs, architecture, music, and handicrafts, along with tasks in classical language understanding such as semantic and syntactic analysis.

\textbf{Data Quality:} While ensuring the authority of the data, we also take specific measures to mitigate data contamination, as detailed in section 3.2.

Our benchmark is organized into 3 major categories and 13 subjects, encompassing general historical knowledge as well as both short- and long-text comprehension of ancient Chinese. In alignment with the methodology proposed by \citet{huang2023ceval}, we adopt a uniform question format, presenting each question with four answer options. Each subject within the benchmark contains an average of over 200 questions, of which five with explanations are designated for development sets. The statistical summary of AC-EVAL is depicted in Table \ref{stat}, and a more detailed statistical breakdown is available in Appendix \ref{a_stat}.

\subsection{Data Collection}

\textbf{Subject Selection:} Our benchmark encompasses general historical knowledge and classical Chinese text comprehension. For the former, we have identified five subcategories, namely: Historical Facts, Geography, Social Customs, Art and Cultural Heritage, Philosophy and Religion. 
For the latter, we distinguish between short texts, which include tasks such as Lexical Pragmatics Analysis, Allusions and Idioms, Word Sense Disambiguation, Translation, and Event Extraction, and long texts, which cover Sentence Pauses, Summarization and Analysis, and Poetry Appreciation, as illustrated in Figure \ref{overview}.

\textbf{Data Source:} The dataset is derived from four main sources: (1) the Complete Library in Four Branches (Siku Quanshu), offering a comprehensive collection of ancient Chinese texts; (2) specialized books on ancient Chinese social customs, architectural history, music history, and geography; (3) official or mock examinations; and (4) existing non-multiple-choice datasets on ancient Chinese, such as GuwenEE.

\textbf{Data processing:} Initially, we recruited undergraduate students and linguistics experts as annotators to manually gather and compile preliminary questions and answers from these sources. The data then underwent a three-fold modification and review process:  
\textbf{(1) Ethical Considerations:} We categorized our data source into reference materials (Sources 1, 2, and 4) and examinations (Source 3). The reference materials were manually adapted to create new questions and answers. Meanwhile, all materials are cited appropriately in Appendix \ref{a_stat}. The examination data, available freely online, were also included. \textbf{(2) Data Contamination:} We aimed to strike a balance between maintaining the authority of the data sources and minimizing data contamination. With the awareness that official examinations might be inadvertently captured and utilized in training LLMs, these were adapted by experts to retain the examinations' core focus while altering the content to some extent. \textbf{(3) Coverage and Accuracy:} We adhered strictly to our design principles to ensure the dataset's diversity and accuracy. A 5\% random sample of the data underwent a quality check, with any found inaccuracies necessitating rework until achieving 100\% accuracy.

Further details on the concepts and data sources for each subject are provided in Appendix \ref{a_stat}.

\subsection{Evaluation}
Accuracy is the primary metric for our evaluation. The ground-truth labels of the development set are public, while the labels of the test set remain confidential to avoid their unintended inclusion in the pre-training corpora. Participants can email to request access to the AC-EVAL under a confidential agreement. Once access is granted, they are encouraged to submit their model predictions via email for evaluation. Results will be provided within one working day and displayed on a public leaderboard on GitHub. This streamlined process ensures secure and fair evaluation.

\begin{figure*}[htbp]
	\centering
	\includegraphics[width=0.96\linewidth]{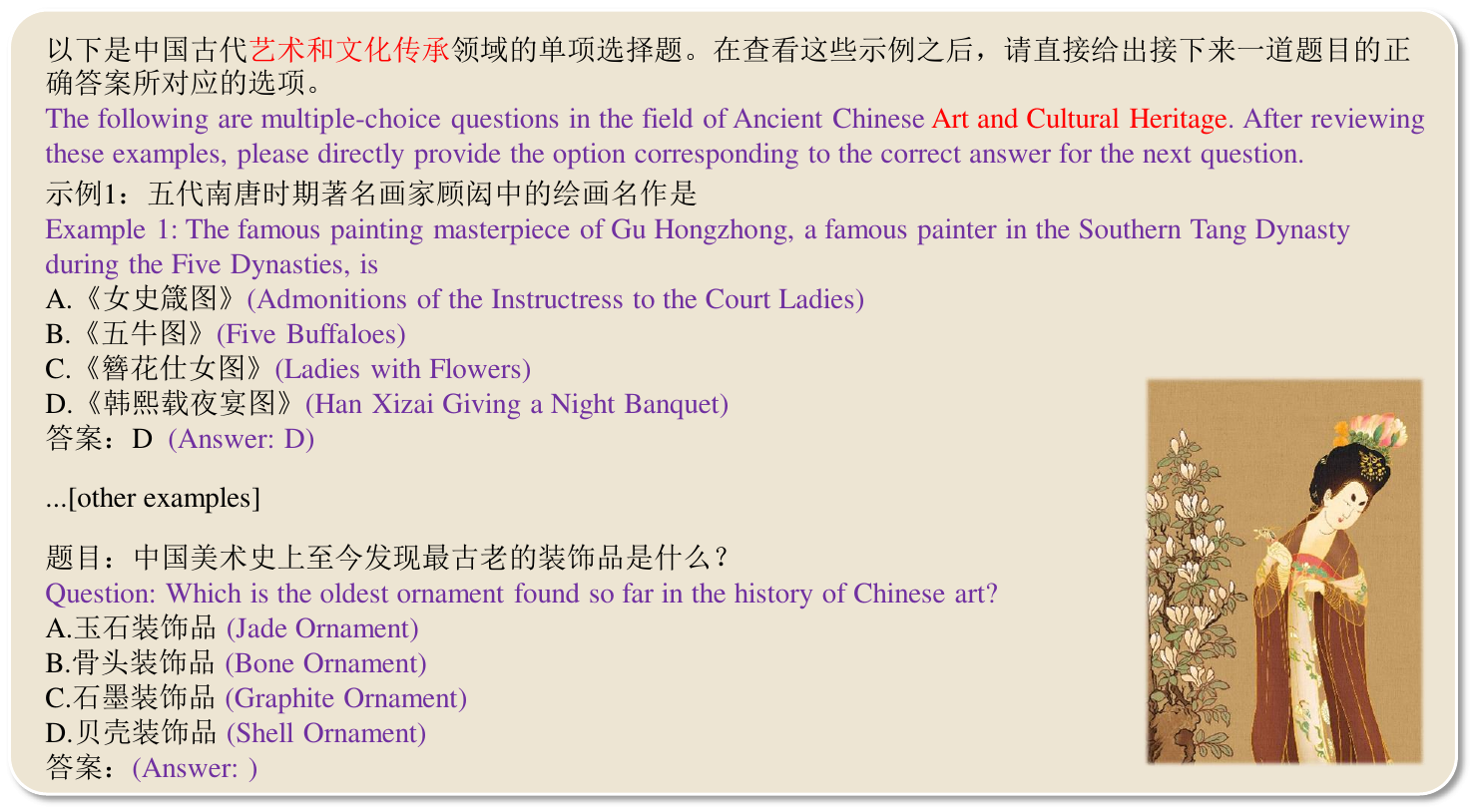}
	\caption{Illustrative few-shot AO prompts from AC-EVAL with corresponding English translations for better readability.}
	\label{prompt}
\end{figure*}

\section{Experiment}
\subsection{Setup}

For evaluation of the AC-EVAL benchmark, we assess LLMs in both zero-shot and few-shot settings, with the few-shot samples drawn from the development set. To extract the answer choices from the models' responses, We employ regular expressions followed by manual verification to ensure successful retrieval in nearly all cases.

We report the results for both  answer-only (AO) and chain-of-thought (COT) \cite{wei2022chain, dong2022survey, zhang2022automatic} settings in zero- and few-shot scenarios. For zero-shot AO setting, we craft prompts in the format: "以下是中国古代[主题]领域的单项选择题，请直接给出正确答案对应的选项。(The following is a multiple-choice question in the field of Ancient Chinese [subject]. Please directly provide the option corresponding to the correct answer.)" For the few-shot AO setting, an example of it prompt is displayed in Figure \ref{prompt}. For the COT settings, their prompts are shown in Appendix \ref{a_prompt}.

Generally, few-shot defaults to five-shot. It is noteworthy that for both the five-shot and five-shot-COT settings, input lengths sometimes surpass the maximum token limit of the models. To accommodate this, we dynamically adjust the number of samples to ensure they fit within the models' context window constraints.

\subsection{Models}
In our evaluation, we select $17$ top-performing LLMs that demonstrate proficiency in Chinese language comprehension. These models represent a variety of organizations and encompass a range of parameter sizes.
For commercial models, we evaluate via API calls, including (1) GPT-4 and GPT-3.5 Turbo \cite{achiam2023gpt}, (2) ERNIE-Bot 4.0 and ERNIE-Bot\footnote{https://cloud.baidu.com/}, (3) GLM-4 and GLM-3-Turbo \cite{zeng2022glm}, (4) Qwen-max \cite{bai2023qwen}. For models with open-sourced parameters, we evaluate (1) LLaMA2-70B \cite{touvron2023llama} (2) Qwen-7B/14B/72B-Chat \cite{bai2023qwen}, (3) Yi-6B/34B-Chat\footnote{https://huggingface.co/01-ai}, (4) Baichuan2-7B/13B-Chat \cite{baichuan2023baichuan2}, (5) ChatGLM3-6B\footnote{https://github.com/THUDM/ChatGLM3}, and (6) Xunzi-Qwen-Chat\footnote{https://github.com/Xunzi-LLM-of-Chinese-classics/XunziALLM}—an LLM that has been continually pre-trained on ancient Chinese corpora based on the Qwen-7B-Chat architecture.
A detailed description of the evaluated models, including their architectural details, pre-training corpora, and versions, is available in Appendix \ref{models_e}. 
We conduct timely evaluations to capture the latest performance levels of these models\footnote{All models were evaluated during 5-10 February 2024.}.

\begin{table*}[h!]
	\centering
	\begin{tabular}{lcccc}
		\hline
		\textbf{Model}              & \textbf{\begin{tabular}[c]{@{}c@{}}General Historical\\Knowledge\end{tabular}} & \textbf{\begin{tabular}[c]{@{}c@{}}Short Text\\Understanding\end{tabular}} & \textbf{\begin{tabular}[c]{@{}c@{}}Long Text\\Understanding\end{tabular}} & \textbf{Average}  \\ \hline
		GPT-4       & 66.11                                                                 & 55.11                                                             & 47.38                                                            & 56.20    \\
		GPT-3.5 Turbo     & 53.50                                                                 & 43.72                                                             & 36.94                                                            & 44.72    \\
		ERNIE-Bot 4.0      & 77.54                                                                 & 68.11                                                             & 66.42                                                            & 70.69    \\
		ERNIE-Bot          & 68.81                                                                 & 57.80                                                             & 51.47                                                            & 59.36    \\
		GLM-4              & 76.63                                                                 & 66.66                                                             & 67.70                                                            & 70.33    \\
		GLM-3-Turbo        & 75.21                                                                 & 60.52                                                             & 59.77                                                            & 65.17    \\
		Qwen-max           & 73.77                                                                 & 64.88                                                             & 63.84                                                            & 67.50    \\ 
		LLaMA-70B			& 33.55                                                                 & 36.29                                                             &       30.72                                                     &   33.54 \\ 
		Qwen-72B-Chat      & 71.25                                                                 & 61.48                                                             & 59.80                                                            & 64.18    \\
		Yi-34B-Chat        & 72.66                                                                 & 61.33                                                             & 58.36                                                            & 64.12    \\
		Qwen-14B-Chat      & 69.51                                                                 & 56.53                                                             & 57.38                                                            & 61.14    \\
		Baichuan2-13B-Chat & 65.57                                                                 & 49.24                                                             & 35.40                                                            & 50.07    \\
		Qwen-7B-Chat       & 62.74                                                                 & 48.76                                                             & 44.97                                                            & 52.16    \\
		Baichuan2-7B-Chat  & 64.38                                                                 & 46.77                                                             & 40.33                                                            & 50.49    \\
		Yi-6B-Chat         & 66.70                                                                 & 47.79                                                             & 39.49                                                            & 51.33    \\
		ChatGLM3-6B        & 58.04                                                                 & 43.01                                                             & 39.73                                                            & 46.93    \\
		Xunzi-Qwen-Chat    & 60.20                                                                 & 44.31                                                             & 30.87                                                            & 45.13   \\ \hline
	\end{tabular}
	\caption{Zero-shot AO average accuracy of all models. We report average accuracy over subjects within each category. “Average” = average over all categories. Models are ranked by model size.}
	\label{zero}
\end{table*}

\section{Results}
In this section, we explore the comparative performance of various models under four distinct settings:  zero-shot AO  as discussed in Section \ref{5.1}, few-shot AO in Section \ref{5.2},  zero- and few-shot COT in Section \ref{5.3}.

\subsection{Zero-shot AO}
\label{5.1}
Given that zero-shot scenarios are among the most common use cases, understanding model performance in this context is crucial. Therefore, we first report the average accuracy in the zero-shot AO setting in Table \ref{zero}, while detailed accuracy breakdowns by subject are provided in Appendix \ref{breakdown}. Our comparison analysis focuses on two critical dimensions: model parameter size and task category.

\textbf{Comparison by model.}
For large models: ERNIE-Bot-4.0 and GLM-4 stand out as top-performing models in ancient Chinese, with accuracies of 70.69\% and 70.33\%, respectively, followed by Qwen-max at 67.50\%. 
Despite primarily being trained on modern Chinese corpora, these LLMs show strong generalization abilities to ancient Chinese. 
For models primarily trained on English corpora, GPT-4 and GPT-3.5 significantly outperform LLaMA-70B. Considering our benchmark is entirely in Chinese, this suggests GPT models' superior generalization capabilities over LLaMA2-70B in handling extensive Chinese content. 
Interestingly, GPT series models perform worse than Chinese LLMs, diverging from conclusions drawn from previous benchmarks in the Chinese domain where GPT often ranked first \cite{li2023cmmlu,huang2023ceval,xu2023superclue}. This indicates that ancient Chinese acts as a low-resource language for English LLMs, highlighting the significant linguistic differences between ancient and modern Chinese. This observation also underscores the importance of our benchmark from another perspective.

For small models: The Yi-34B-Chat showcases remarkable parameter efficiency and performs comparably to larger models like Qwen-72B-Chat. This efficiency can be attributed to their extensive training on large-scale Chinese corpora and architectural optimizations. Qwen-14B even surpasses the GPT series and ERNIE-bot, presenting high cost-effectiveness in ancient Chinese comprehension as a relatively smaller open-source model. Qwen-7B-Chat achieves the best performance among models with less than 10B parameters. Baichuan2-13B-Chat, compared to Baichuan2-7B-Chat, does not show performance improvement despite increased parameters, possibly due to a reduction in focus on ancient Chinese content in its training corpus. Xunzi-Qwen-Chat, despite being fine-tuned on ancient Chinese texts, shows a decline in performance compared to Qwen-7B-Chat. This highlights the trade-off between specialized knowledge and general applicability.

\begin{table*}[h!]
	\centering
	\begin{tabular}{lllll}
		\hline
		\textbf{Model} & \textbf{\begin{tabular}[c]{@{}c@{}}General Historical\\ Knowledge\end{tabular}} & \textbf{\begin{tabular}[c]{@{}c@{}}Short Text\\ Understanding\end{tabular}} & \textbf{\begin{tabular}[c]{@{}c@{}}Long Text\\ Understanding\end{tabular}} & \textbf{Average} \\ \hline
		GPT-4          & 65.91 {\color{gray}(-0.20)} &               58.07 (+2.96) &               48.36 (+0.98) &               57.45 (+1.25)       \\ 
		GPT-3.5 Turbo        &  53.99 (+0.49) &  43.21 {\color{gray}(-0.51)} &  36.40 {\color{gray}(-0.54)} &  44.54 {\color{gray}(-0.18)}     \\ 
		ERNIE-Bot 4.0        & 75.69 {\color{gray}(-1.85)} &               69.59 (+1.48) &  66.12 {\color{gray}(-0.30)} &  70.47 {\color{gray}(-0.22)}          \\ 
		ERNIE-Bot        &  68.81 (+0.00) &  57.62 {\color{gray}(-0.18)} &  50.36 {\color{gray}(-1.11)} &  58.93 {\color{gray}(-0.43)}        \\ 
		GLM-4        & 74.89 {\color{gray}(-1.74)} &  65.48 {\color{gray}(-1.18)} &               69.07 (+1.37) &  69.81 {\color{gray}(-0.52)}      \\ 
		GLM-3-Turbo        & 72.99 {\color{gray}(-2.22)} &  59.48 {\color{gray}(-1.04)} &  59.66 {\color{gray}(-0.11)} &  64.04 {\color{gray}(-1.13)}    \\ 
		Qwen-max     &  75.29 (+1.52) &               65.48 (+0.60) &               66.99 (+3.15) &               69.25 (+1.75)    \\
		Qwen-72B-Chat        & 71.67 (+0.42) &  61.30 {\color{gray}(-0.18)} &  57.07 {\color{gray}(-2.73)} &  63.35 {\color{gray}(-0.83)}          \\ 
		Yi-34B-Chat        &  66.62 {\color{gray}(-6.04)} &  52.57 {\color{gray}(-8.76)} & 41.90 {\color{gray}(-16.46)} & 53.70 {\color{gray}(-10.42)}     \\ 
		Qwen-14B-Chat        &  70.60 (+1.09) &  53.73 {\color{gray}(-2.80)} & 45.91 {\color{gray}(-11.47)} &  56.75 {\color{gray}(-4.39)}       \\ 
		Baichuan2-13B-Chat       & 63.75 {\color{gray}(-1.82)} &  45.86 {\color{gray}(-3.38)} &  32.74 {\color{gray}(-2.66)} &  47.45 {\color{gray}(-2.62)}   \\ 
		Qwen-7B-Chat      & 61.42 {\color{gray}(-1.32)} &  45.98 {\color{gray}(-2.78)} & 30.78 {\color{gray}(-14.19)} &  46.06 {\color{gray}(-6.10)}     \\ 
		Baichuan2-7B-Chat & 63.37 {\color{gray}(-1.01)} &  45.91 {\color{gray}(-0.86)} &  39.94 {\color{gray}(-0.39)} &  49.74 {\color{gray}(-0.75)}         \\ 
		Yi-6B-Chat & 55.76 {\color{gray}(-10.94)} & 35.97 {\color{gray}(-11.82)} & 28.48 {\color{gray}(-11.01)} & 40.07 {\color{gray}(-11.26)}      \\ 
		ChatGLM3-6B     & 55.74 {\color{gray}(-2.30)} &  42.92 {\color{gray}(-0.09)} &  38.45 {\color{gray}(-1.28)} &  45.71 {\color{gray}(-1.22)}          \\ 
		Xunzi-Qwen-Chat    & 51.30 {\color{gray}(-8.90)} &  41.25 {\color{gray}(-3.06)} &  29.84 {\color{gray}(-1.03)} &  40.80 {\color{gray}(-4.33)}           \\ \hline
	\end{tabular}
	\caption{Few-shot AO average accuracy of all models. We report average accuracy over subjects within each category. "Average" = average over all categories. The values in parentheses show the relative change compared to the zero-shot AO scenario.}
	\label{five}
\end{table*}

\textbf{Comparison by Task Category.}
(1) General Historical Knowledge: Most models score highest on this category of tasks, likely because these tasks focus on the retrieval and understanding of factual information without necessitating deep textual analysis and reasoning.
(2) Short Text Understanding: Compared to long text comprehension, models generally score higher on short text understanding, though still lower than on general historical knowledge tasks. This may be because short text understanding still requires models to capture subtle semantic differences and contextual relationships, albeit with relatively lower complexity.
(3) Long Text Comprehension: All models generally score lower on long text comprehension than on other tasks, indicating it as a challenging task that requires advanced understanding, reasoning, and synthesis capabilities.

\subsection{Few-shot AO}
\label{5.2}

Table \ref{five} presents the results for the few-shot AO setting, alongside a comparison with the zero-shot AO setting. 

\textbf{For large models}, only GPT-4 and Qwen have a 1-2\% improvement in this setting, while others slightly declined, diverging from previous Chinese benchmarks where few-shot usually excelled over zero-shot \cite{huang2023ceval}. We attribute this discrepancy primarily to the task specificity. 
Previous benchmarks often encompassed a broader range of subjects and task categories, including scientific, technological, and coding tasks. In such tasks, few-shot learning effectively aids models in capturing underlying patterns, thereby enhancing adaptability and generalization.  However, our tasks focus on a broad spectrum of fragmented knowledge and require a deep understanding of ancient Chinese, including its cultural, historical backgrounds, and linguistic structures, leading to a unique challenge where few-shot learning might not provide the same level of benefit. Instead, the specificity and complexity of ancient Chinese tasks can result in the introduction of noise or irrelevant information through few-shot examples, potentially hindering model performance rather than enhancing it.

\textbf{For small models}, the decline in performance is more pronounced. The Yi-series models, although performing comparably to large models in the zero-shot AO scenario, show the most significant drop in the few-shot setting, with a performance decrease of 10\%. This aligns with conclusions from some previous studies \cite{li2023cmmlu}, suggesting that for smaller models, few-shot learning may introduce too much irrelevant content, potentially leading to information interference. In such instances, models might struggle to extract useful knowledge or patterns from a few samples, as the additional information introduced may not be entirely relevant to the task, thereby diluting the model's focus.

From this analysis, we conclude that despite the task specificity causing few-shot learning to sometimes act as interference, large models possess stronger language understanding capabilities and higher stability in processing distracting information, even achieving improvements on some tasks. However, smaller models struggle due to insufficient parameters to effectively encode and utilize contextual information.

\begin{figure}[t]
	\centering
	\includegraphics[width=1\linewidth]{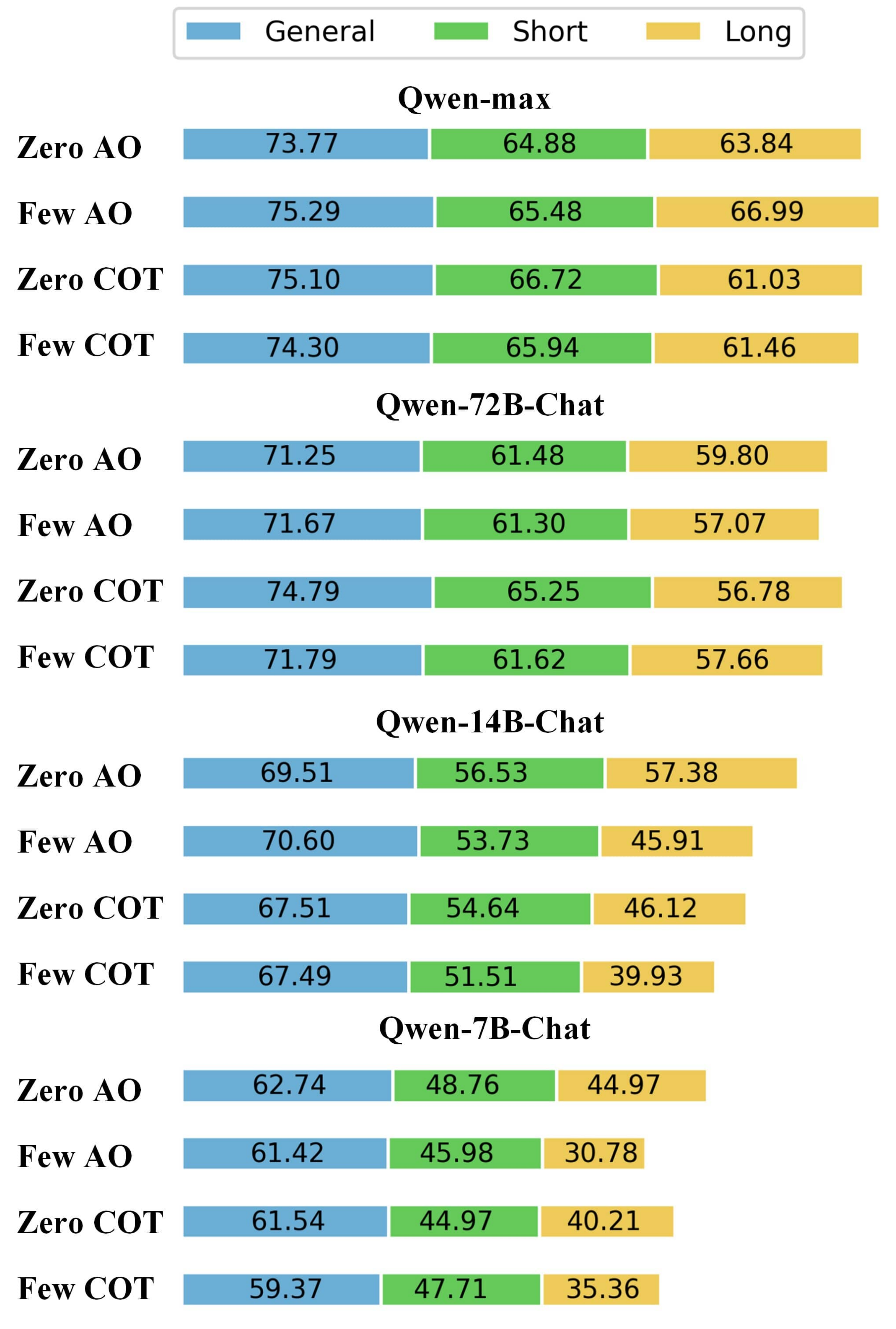}
	\caption{Average accuracy for each category in various settings with different model sizes. Here, we have omitted the category names and shot types for brevity.}
	\label{COT}
\end{figure}

\subsection{Chain-of-thought}
\label{5.3}

As shown in Figure \ref{COT}, we conduct a series of experiments to explore the impact of COT on LLMs of varying parameter sizes. Our experimental setup includes two scenarios: zero-shot COT and few-shot COT (prompts are detailed in Appendix \ref{a_prompt}). To facilitate a nuanced comparison, we select the Qwen series, encompassing four different sizes: 7B, 14B, 72B, and the non-public Qwen-max. Given that few-shot performance does not surpass zero-shot on AC-EVAL, our analysis primarily contrasts the two COT formats against the zero-shot AO scenario.

\textbf{Zero-shot COT vs. Zero-shot AO:}
In zero-shot COT, prompts are adjusted to encourage stepwise analysis. This method particularly benefits large models like Qwen-max and Qwen-72B-Chat in historical knowledge and short text understanding tasks but shows a decrease in long text comprehension. 
We attribute this to the increased reasoning steps needed for long text understanding in large models, where any small errors can accumulate, negatively impacting the final answer's accuracy.

As model size decreases, a downward trend in performance is evident across all tasks in the zero-shot COT setting. This decline is likely due to the COT method's demand for models to understand the question, generate intermediate reasoning steps, and ultimately formulate an answer. This process, more complex than direct answer generation, requires robust semantic understanding and logical reasoning capabilities. With reduced model parameters, the capability of models to perform these functions weakens, leading to diminished performance.

\textbf{Few-shot COT vs. Zero-shot AO/COT:}
Few-shot COT underperforms in comparison to zero-shot settings in both AO and COT across all parameter sizes. This aligns with our above observation that few-shot learning generally offers less benefit in our benchmark, which demands a broad understanding of fragmented knowledge and deep comprehension of ancient Chinese, including its cultural, historical backgrounds, and linguistic structures. The unique challenges posed by these requirements suggest that even structured COT, when combined with few-shot examples, may be perceived as informational noise, thereby impeding the model's ultimate reasoning capability.

Through this analysis, it is evident that while COT reasoning can enhance model performance in certain contexts, the effectiveness of this approach is contingent upon the model's capacity for complex information processing and logical deduction. The decline in performance with reduced model size and the limited impact of few-shot learning highlight the intricate balance required between model abilities, task specificity, and the introduced format of COT.

\section{Conclusion}

We introduce AC-EVAL, a benchmark designed to evaluate LLMs' proficiency in ancient Chinese, addressing a gap by covering historical knowledge and language understanding extensively. Our experiments reveal significant improvement areas for existing LLMs. We identify critical factors influencing LLM performance and suggest practical directions for enhancing these models. AC-EVAL aims to advance LLM application in ancient Chinese education, offering a valuable tool for assessing and developing Chinese LLMs.

\section{Limitations}
While our study introduces the AC-EVAL benchmark as a robust tool for evaluating LLMs in the domain of ancient Chinese, it is imperative to acknowledge several limitations that accompany our research:

\textbf{Absence of Human Baseline:} The lack of a human comparative standard impedes the evaluation of LLMs' depth of understanding, cultural acuity, and contextual sensitivity relative to the insights provided by scholars specializing in ancient Chinese literature. Consequently, while the AC-EVAL benchmark may offer quantitative evaluations of LLM proficiency, it might not capture the qualitative dimensions of linguistic and cultural comprehension that are crucial in the analysis of ancient Chinese texts.

\textbf{Focus on Multiple-Choice Questions:} The current iteration of AC-EVAL primarily utilizes a multiple-choice format to assess LLMs. This approach, while effective in certain assessments, does not measure the generative capabilities of LLMs. For example, poetry generation \cite{chen2019sentiment,zhipeng2019jiuge}. As a result, our benchmark may not fully capture the models' ability to produce coherent and contextually relevant responses in an open-ended format.

In light of these limitations, future work will aim to incorporate human evaluation and expand the benchmark to include open-ended and generative tasks, thereby enhancing the comprehensive assessment of models' capabilities.

\section*{Acknowledgments}
This work is supported by the National Natural Science Foundations of China under Grant (61972047, 62372060), and the NSFC-General Technology Basic Research Joint Funds under Grant (U1936220).

\bibliography{custom}
 
\appendix

\section{Details of AC-EVAL}
\label{a_stat}
Table \ref{source and concept} provides a comprehensive overview of the AC-EVAL, detailing the data sources and the specific concepts addressed within each subject. Table \ref{num and avg} offers insights into the quantitative aspects of the dataset, including the number of questions per subject and their average length (accounting for both the questions and explanations, measured in characters). Furthermore, Table \ref{ABCD} shows the distribution of choices across the multiple-choice questions.

\begin{table}[htbp]
	\centering
	\begin{tabular}{ccc}
		\hline
		Option & C-EVAL & AC-EVAL     \\ \hline
		A & 22.9\%  & 26.2\%  \\
		B & 26.0\%  & 26.5\%  \\
		C & 26.4\%  & 23.6\%  \\
		D & 24.7\%  & 23.7\%  \\ \hline
	\end{tabular}
	\caption{Distribution of Answers}
	\label{ABCD}
\end{table}

\section{Details of Annotation}
\begin{figure*}[htbp]
	\centering
	\includegraphics[width=1\linewidth]{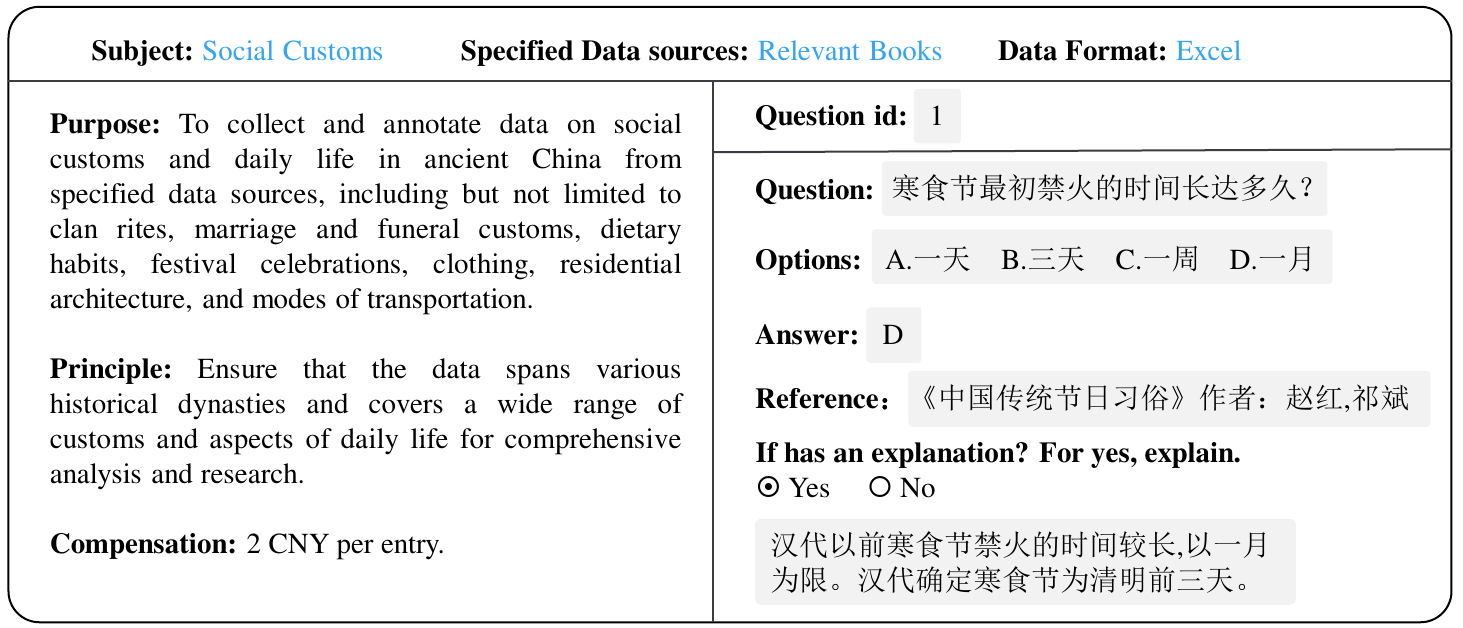}
	\caption{Illustration of the Annotation Process: An Example of Social Customs Data Annotation}
	\label{annotation}
\end{figure*}

Figure \ref{annotation} details the specific requirements for data annotation. All our annotators are Chinese undergraduate students and experts in Chinese linguistics, and they are compensated at a rate that meets market standards.

\section{Prompts for Evaluation}
\label{a_prompt}
\begin{figure*}[htbp]
	\centering
	\includegraphics[width=1\linewidth]{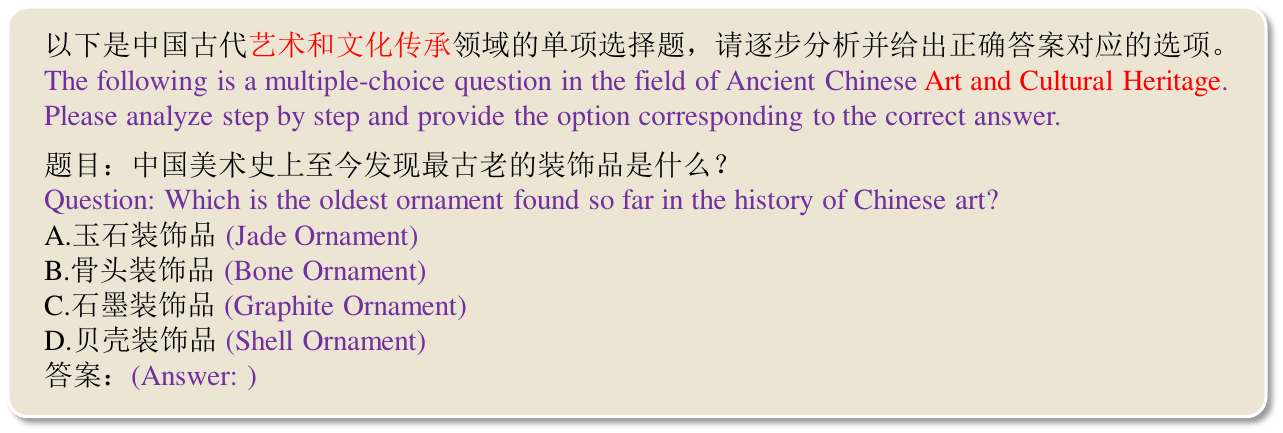}
	\caption{Illustrative zero-shot COT prompts from AC-EVAL with corresponding English translations for better readability.}
	\label{zero_COT}
\end{figure*}

\begin{figure*}[htbp]
	\centering
	\includegraphics[width=1\linewidth]{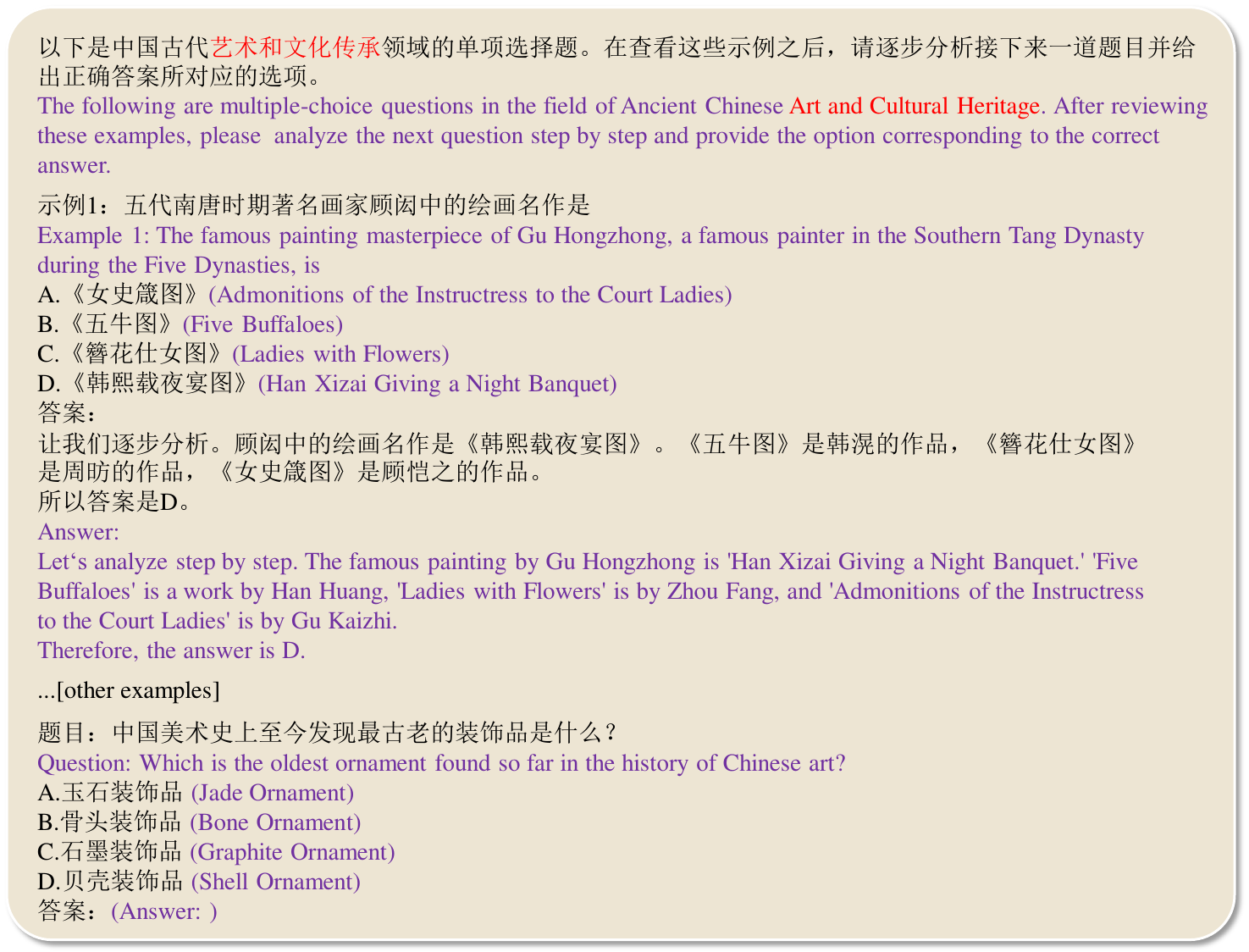}
	\caption{Illustrative few-shot COT prompts from AC-EVAL with corresponding English translations for better readability.}
	\label{few_COT}
\end{figure*}

Figures \ref{zero_COT} and \ref{few_COT} display the chain-of-thought evaluation prompts used in the zero-shot and few-shot settings, respectively.

\section{Details of the LLMs being evaluated}
\label{models_e}

We provide a detailed description of the Large Language Models (LLMs) that were rigorously evaluated during the period of 5-10 February 2024, ensuring the assessment of the latest model versions prior to submission.

\textbf{GPT} \cite{achiam2023gpt} series models, developed by OpenAI, designed to be more aligned with human-like interaction, exhibiting helpful, safe, and truthful behavior as enhanced by Reinforcement Learning from Human Feedback (RLHF). GPT-4, with its ability to process images, PDFs, and other file types, underwent a comprehensive post-training alignment process. We evaluate the versions of gpt-3.5-turbo-0125 and gpt-4-0125-preview.

\textbf{ERNIE-Bot} is an industrial-grade, knowledge-enhanced LLM developed by Baidu. The 4.0 version represents a significant upgrade in understanding, generation, logic, and memory capabilities over its predecessors, supporting extensive input and output lengths (5K input + 2K output). Our evaluation included both ERNIE-Bot and ERNIE-Bot 4.0.

\textbf{GLM} \cite{zeng2022glm} series, developed by Zhipu AI and Tsinghua University, are bidirectional dense models excelling in bilingual language processing. ChatGLM, a derivative of GLM, targets Chinese QA and dialogue tasks with enhanced fine-tuning and feedback. We evaluate ChatGLM3-6B and the commercial GLM-3-Turbo and GLM-4.

\textbf{Qwen} \cite{bai2023qwen}, developed by Alibaba, is trained on a vast corpus including 3 trillion tokens of texts and codes. The chat variants of Qwen have been refined through RLHF to better align with human preferences. 
We conduct a comprehensive evaluation of the Qwen series, covering multiple versions with varying parameter sizes, including Qwen-7B/14B/72B-Chat and Qwen-max.

\textbf{LLaMA2} \cite{touvron2023llama}, developed and open-sourced by Meta AI, excels in encoding, inference, and knowledge application. It incorporates several enhancements over the vanilla Transformer architecture employed by preceding LLMs, optimizing for greater training efficiency. In our experiment, we evaluate the performance of the LLaMA2-70B version.

\textbf{Yi} series models by 01.AI are open-source bilingual models trained from scratch on a 3T multilingual corpus, featuring an extended context window of up to 200K tokens. We utilize the Yi-6B-Chat and Yi-34B-Chat versions, which support up to 32K tokens for context in inferences.

\textbf{Baichuan2} \cite{baichuan2023baichuan2} is developed by Baichuan Intelligence Inc., trained on a 2.6 trillion token high-quality corpus and supporting multiple languages including Chinese, English and others. The versions evaluated are Baichuan2-7B-Chat and Baichuan2-13B-Chat.

\textbf{Xunzi} is a model collaboratively released by Nanjing Agricultural University and the Zhonghua Book Company. It is fine-tuned on classical Chinese corpora such as the Siku Quanshu, based on foundations from Qwen, Baichuan, and GLM. We evaluate the Xunzi-Qwen-Chat, a model trained from Qwen-7B-Chat.

\section{Breakdown of Model Performance}
\label{breakdown}
Table \ref{zero_five_AO}  provides a detailed accuracy breakdown by subject for four representative models under AO settings in both zero- and few-shot scenarios, respectively. Comprehensive results for all models are made available on GitHub.

\begin{table*}[htbp]
	\centering
	\scalebox{0.8}{
		\begin{tabular}{lll}
			\hline \rule{0pt}{10pt}
			\textbf{Subject} & \textbf{Data Source} & \textbf{Concepts} 
			\\ \hline \rule{0pt}{30pt}
			Historical Facts & Official history exams & \begin{tabular}[l]{@{}l@{}} Historical facts, covering political, \\economic, and military developments\\ across different periods. \end{tabular} 
			\\ \rule{0pt}{30pt}
			Geography & \begin{tabular}[l]{@{}l@{}} 10\% from mock exams and \\90\% from ancient place \\ names knowledge database \end{tabular} & \begin{tabular}[l]{@{}l@{}}Administrative divisions, historical\\ boundaries,changes in place \\names over time. \end{tabular} 
			\\ \rule{0pt}{40pt}
			Social Customs & \begin{tabular}[l]{@{}l@{}}Relevant Books , e.g., \\Customs of the Qing Dynasty. \end{tabular} & \begin{tabular}[l]{@{}l@{}}Changes in clothing, food, housing, \\transportation, traditional festivals,\\weddings and funerals, family etiquette,\\ public customs, business practices,\\entertainment customs over time. \end{tabular} 
			\\ \rule{0pt}{30pt}
			Art and Cultural Heritage & \begin{tabular}[l]{@{}l@{}}Mock exams and relevant books,\\e.g., History of Chinese Art, \\Architecture and Music History \end{tabular} & \begin{tabular}[l]{@{}l@{}}Changes in calligraphy, painting, \\architecture, craftsmanship and music \\over different periods. \end{tabular} 
			\\ \rule{0pt}{30pt}
			Philosophy and Religion & \begin{tabular}[l]{@{}l@{}}Mock exams and relevant books \\e.g.,  History of Chinese philosophy. \end{tabular} & \begin{tabular}[l]{@{}l@{}}Changes in the content of Taoism, \\ Confucianism, Buddhism, etc., and  \\their rise and decline over time. \end{tabular} \\
			\hline \rule{0pt}{20pt}
			Lexical Pragmatic Analysis & Compiled by linguistics experts & \begin{tabular}[l]{@{}l@{}}Flexible usage of parts of speech \\ and figures of speech.  \end{tabular} 
			\\ \rule{0pt}{20pt}
			Allusions and Idioms & Official and mock exam questions & \begin{tabular}[l]{@{}l@{}}Allusions and idioms and the cultural \\ meaning behind them. \end{tabular} 
			\\ \rule{0pt}{20pt}
			Word Sense Disambiguation & \begin{tabular}[l]{@{}l@{}}Word Sense Disambiguation \\Dataset \cite{shu-etal-2021-gu} \end{tabular}& \begin{tabular}[l]{@{}l@{}}Explanation of word meaning in\\ a given text.\end{tabular} 
			\\ \rule{0pt}{20pt}
			Translation & \begin{tabular}[l]{@{}l@{}}Classical-Modern Chinese \\ translation dataset\footnote{https://github.com/NiuTrans/Classical-Modern} \end{tabular} & \begin{tabular}[l]{@{}l@{}}Overall understanding of the\\semantics and syntax of sentences. \end{tabular}
			\\ \rule{0pt}{30pt}
			Event Extraction & GuwenEE\footnote{https://github.com/Lyn4ever29/GuwenEE} & \begin{tabular}[l]{@{}l@{}}Identifying basic facts and information \\in short texts, such as time, location, \\characters, event types, etc. \end{tabular}
			\\ \hline \rule{0pt}{20pt}
			Sentence Pauses & Siku Quanshu & \begin{tabular}[l]{@{}l@{}}Make pauses in reading unpunctuated \\ancient writings. \end{tabular} 
			\\ \rule{0pt}{20pt}
			Summarization and Analysis & Official and mock exams & \begin{tabular}[l]{@{}l@{}}Overall understanding, analysis, and \\ reasoning for classical Chinese texts \end{tabular}
			\\ \rule{0pt}{20pt}
			Poetry Appreciation & Official and mock exams & \begin{tabular}[l]{@{}l@{}}Analysis of imagery, style, sentiment\\in classical Chinese poetry \end{tabular}  \\
			\hline
	\end{tabular}}
	\caption{Data Sources and Concepts for All Subjects.}
	\label{source and concept}
\end{table*}

\begin{table*}[htbp]
	\centering
	
	\begin{tabular}{lccccc}
		\hline
		\multirow{2}{*}{Subject} & \multicolumn{2}{c}{Test} & \multicolumn{3}{c}{Dev}  \\ \cline{2-6}
		& \# Questions & Len. of $Q$ & \# Questions &  Len. of $Q$ &  Len. of $E$ \\ \hline
		Historical Facts    & 199      & 157.1     & 5       & 138.0   & 200.2 \\
		Geography    & 197      & 33.8      & 5       & 32.8  &  33.6  \\
		Social Customs & 202      & 48.5      & 5       & 48.6   &  65.0 \\
		Art and Cultural Heritage   & 195      & 35.8      & 5       & 32.4   &  56.8   \\
		Philosophy and Religion     & 196      & 39.2      & 5       & 48.0   &  77.4   \\ \hline
		Lexical Pragmatic Analysis     & 198      & 62.5     & 5       & 69.6  & 75.4 \\
		Allusions and Idioms   & 206      & 191.2     & 5       & 79.6 &  132.4  \\
		Word Sense Disambiguation      & 402      & 176.6     & 5       & 163.2  &  91.4 \\
		Translation     & 199      & 409.1     & 5       & 315.0   &  79.4 \\  
		Event Extraction      & 185      & 238.8    & 5       & 150.4  & 109.0 \\ \hline
		Sentence Pauses     & 202      & 390.2     & 5       & 404.2  & 294.2  \\
		Summarization and Analysis  & 598      & 880.5     & 5       & 856.0  &  341.4  \\
		Poetry Appreciation      & 201      & 339.4     & 5       & 371.8  & 109.0 \\ \hline
	\end{tabular}
	\caption{Quantitative Statistics for All Subjects.}
	\label{num and avg}
\end{table*}

\begin{table*}[htbp]
	\begin{tabular}{lcccc}
		\hline
		\multicolumn{1}{c}{\textbf{Subject}} & \multicolumn{1}{c}{\textbf{ERNIE-Bot 4.0}} & \multicolumn{1}{c}{\textbf{GLM-4}} & \multicolumn{1}{c}{\textbf{Yi-34B-Chat}} & \multicolumn{1}{c}{\textbf{Qwen-7B-Chat}} \\ \hline
		\textbf{Historical Facts}            & 78.39/76.76                                & 78.89/84.42                        & 75.88/58.29                              & 61.81/61.31                               \\
		\textbf{Geography}                   & 78.17/74.11                                & 75.13/75.13                        & 71.57/75.63                              & 66.50/62.44                               \\
		\textbf{Social Customs}              & 79.21/78.22                                & 77.23/75.74                        & 76.73/70.79                              & 70.79/71.29                               \\
		\textbf{Art and Cultural Heritage}   & 74.87/72.82                                & 76.92/75.90                        & 70.26/68.21                              & 59.49/61.03                               \\
		\textbf{Philosophy and Religion}    & 77.04/76.53                                & 75.00/63.27                        & 68.88/60.20                              & 55.10/51.02                               \\  \hline
		\textbf{Lexical Pragmatics Analysis} & 78.22/83.17                                & 75.25/79.70                        & 77.72/57.43                              & 44.55/27.23                               \\
		\textbf{Allusions and Idioms}        & 66.67/70.20                                & 58.59/59.60                        & 48.48/44.95                              & 33.84/32.83                               \\
		\textbf{Word Sense Disambiguation}   & 67.96/70.39                                & 69.90/68.93                        & 70.39/64.08                              & 59.71/52.43                               \\
		\textbf{Translation}                 & 56.22/56.22                                & 59.45/52.49                        & 49.50/35.82                              & 39.80/36.57                               \\
		\textbf{Event Extraction}            & 71.86/74.37                                & 71.86/72.86                        & 65.33/47.74                              & 47.74/43.22                               \\  \hline
		\textbf{Sentence Pauses}             & 56.86/52.01                                & 59.20/59.36                        & 42.64/32.44                              & 37.63/31.77                               \\
		\textbf{Summarization and Analysis}  & 64.18/63.18                                & 68.66/68.16                        & 54.73/35.82                              & 52.74/33.33                               \\
		\textbf{Poetry Appreciation}         & 77.84/76.76                                & 73.51/73.51                        & 72.97/70.27                              & 62.70/64.86                              \\ \hline
	\end{tabular}
	\caption{Accuracy per subject in the answer-only setting: a comparison of zero-shot (left) and few-shot (right) performance.}
	\label{zero_five_AO}
\end{table*}

\end{CJK}
\end{document}